\PassOptionsToPackage{table}{xcolor}
\documentclass[10pt,twocolumn,letterpaper]{article}

\usepackage[arxiv]{cvpr}      

\definecolor{cvprblue}{rgb}{0.21,0.49,0.74}
\usepackage[pagebackref,breaklinks,colorlinks,allcolors=cvprblue]{hyperref}
\usepackage{multirow}
\usepackage{algorithmic}
\usepackage{algorithm}
\usepackage{listings}
\usepackage{float}
\usepackage{rotating}

\def\paperID{5015} 
\def\confName{ICCV}
\def\confYear{2025}

\lstdefinestyle{mypython}{
    language=Python,
    basicstyle=\ttfamily\scriptsize,
    keywordstyle=\color{blue},
    stringstyle=\color{red},
    commentstyle=\color{green!70!black},
    backgroundcolor=\color{gray!5},
    showstringspaces=false
}

\def\RETAKE{R\scalebox{0.8}{E}T\scalebox{0.8}{A}K\scalebox{0.8}{E}}

\title{{\RETAKE}: Reducing Temporal and Knowledge Redundancy \\ for Long Video Understanding}

\author{Xiao Wang \\
Harbin Institute of Technology, Shenzhen \\
{\tt\small scz.wangxiao@gmail.com}
\and
Qingyi Si\\
Huawei Technologies Co., Ltd. \\
{\tt\small siqingyi@iie.ac.cn}
\and
Jianlong Wu \\
Harbin Institute of Technology, Shenzhen \\
{\tt\small wujianlong@hit.edu.cn}
\and
Shiyu Zhu \\
Shandong University \\
{\tt\small }
\and
Li Cao \\
Huawei Technologies Co., Ltd.\\
{\tt\small wulin50@huawei.com}
\and
Liqiang Nie \\
Harbin Institute of Technology, Shenzhen \\
{\tt\small nieliqiang@gmail.com}
}
\author{
Xiao Wang\textsuperscript{1}\footnotemark[1] \  \footnotemark[2]  \quad
Qingyi Si\textsuperscript{2}\footnotemark[1] \quad
Jianlong Wu\textsuperscript{1}\footnotemark[3]   \quad
Shiyu Zhu\textsuperscript{3} \quad
Li Cao\textsuperscript{2} \quad
Liqiang Nie\textsuperscript{1}\footnotemark[3]\\
\textsuperscript{1}Harbin Institute of Technology, Shenzhen \\
\textsuperscript{2}Huawei Technologies Co., Ltd.
\textsuperscript{3}Shandong University \\
{\tt\small scz.wangxiao@gmail.com, siqingyi@huawei.com, wujianlong@hit.edu.cn} \\
{\tt\small xyzcaoli@outlook.com, nieliqiang@gmail.com}
}

\begin{document}
\maketitle

\renewcommand{\thefootnote}{\fnsymbol{footnote}}  
\footnotetext[1]{Equal contribution.}  
\footnotetext[2]{Work done during an internship at Huawei.}
\footnotetext[3]{Corresponding author.}

\begin{abstract}

Video Large Language Models (VideoLLMs) have made significant strides in video understanding but struggle with long videos due to the limitations of their backbone LLMs. Existing solutions rely on length extrapolation, which is memory-constrained, or visual token compression, which primarily leverages low-level temporal redundancy while overlooking the more effective high-level knowledge redundancy.
To address this, we propose \textbf{\RETAKE}, a training-free method with two novel modules \textbf{DPSelect} and \textbf{PivotKV}, to jointly reduce both temporal visual redundancy and knowledge redundancy for video compression. 
To align with the way of human temporal perception, DPSelect identifies keyframes based on inter-frame distance peaks. 
To leverage LLMs' learned prior knowledge, PivotKV marks the keyframes as pivots and compresses non-pivot frames by pruning low-attention tokens in their KV cache.  
{\RETAKE} can be plug-and-play adapted to the existing VideoLLMs and 
enables them to process 8× longer frames (up to 2048), outperforming similar-sized models by 3–5\% and even rivaling much larger ones on VideoMME, MLVU, LongVideoBench, and LVBench.
Moreover, by overlapping compression operations with prefilling, {\RETAKE} introduces only ~10\% prefilling latency overhead while reducing decoding latency by ~20\%.
Our code is available at \url{https://github.com/SCZwangxiao/video-ReTaKe}.

\end{abstract}
\section{Introduction}
\label{sec:intro}

In pursuit of general intelligence, Video Large Language Models (VideoLLMs)~\cite{2023videochat, lin2023videollava, zhang2023video, maaz2023video} have revolutionized video understanding. 
However, as a nature extension of Multimodal Large Language Models (MLLMs), they require hundreds of tokens to represent a single image \cite{wang_qwen2-vl_2024, li_llava-onevision_2024}, making it impractical to process videos longer than several minutes. Specifically, the context length of common MLLMs \cite{wang_qwen2-vl_2024, li_llava-onevision_2024} limits their ability to handle only short videos of around 4 minutes at 240 frames.

\begin{figure}[t]
    \centering
    \includegraphics[width=\linewidth]{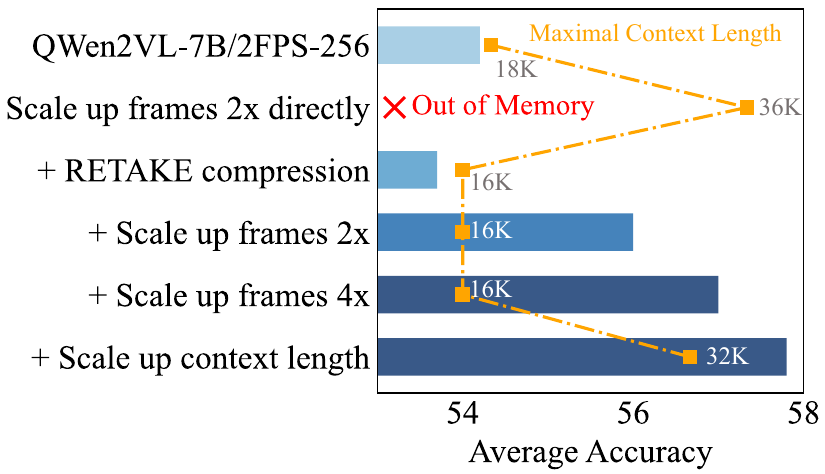}
    \caption{
    \RETAKE\ effectively compresses video sequence in VideoLLMs, allowing longer perception and improved performance within a fixed memory budget (measured by context length).
    }
    \label{fig:frame_scaling_retake}
\end{figure}

To empower VideoLLMs for longer videos, 
\citeauthor{lin2023videollava}~\cite{lin2023videollava} and  \citeauthor{Li_VideoChat2_2024}~\cite{Li_VideoChat2_2024} simply adopt sparse sampling. This approach overlooks the intra-frame spatial details and the inter-frame temporal structure~\cite{weng_longvlm_2024}.
To enable dense frame sampling, some methods \cite{zhang_longva_2024, shu_video-xl_2024} extend the context length of MLLMs through long context post-training. Although they increase performance under long frames, the maximal number of frames is still limited by GPU memory.
Therefore, recent methods \cite{song2023moviechat, jin_chat-univi_2023, he_ma-lmm_2024, weng_longvlm_2024} compress visual tokens to fit more frames into VideoLLM within existing GPU memory limits, including token sparsification both between and within frames \cite{jin_chat-univi_2023} or using visual token compressors like Q-Former \cite{2023videochat}. However, these methods mainly focus on spatial and temporal visual redundancies \cite{wang_rtq_2023, zou_longvideo_review_2024}, which are low-level with high information loss.

A promising solution to these limitations is \textbf{R\scalebox{0.8}{E}}ducing \textbf{T}empor\textbf{\scalebox{0.8}{A}}l and \textbf{K}nowledge r\textbf{\scalebox{0.8}{E}}dundancy (\textbf{\RETAKE}) for compression.
Contrary to low-level temporal redundancy, knowledge redundancy captures high-level abstract patterns or structures that can be inferred from human prior knowledge~\cite{gonzales1987digital_image_process}, allowing for better compression ratios with smaller loss.
The biggest knowledge base in VideoLLMs is the LLM. The attention patterns across LLM layers inherently capture token redundancy, allowing lower-scoring tokens to be dropped without significantly impacting performance~\cite{zhang_h2o_2023}.
In VideoLLMs, the biggest knowledge base is the LLM itself. The attention patterns across its layers naturally capture token redundancy, allowing lower-scoring tokens to be dropped without significant performance impact~\cite{zhang_h2o_2023}.
However, compression based on knowledge redundancy introduces more computational overhead. To mitigate this, {\RETAKE} integrates both knowledge and low-level temporal redundancy, allowing VideoLLMs to process significantly more frames within fixed GPU memory constraints and litter computational overhead, thereby enhancing long-video understanding, as illustrated in \autoref{fig:frame_scaling_retake}\footnote{2FPS-256 means uniform frame sampling at 2FPS, with a limit 256 frames. For longer videos, FPS is reduced to maintain this limit.}.

{\RETAKE} framework comprises two training-free components: \textbf{DPSelect} and \textbf{PivotKV}. 
To reduce low-level temporal redundancy, DPSelect extracts keyframes before feeding into LLM. Based on methods \cite{he_ma-lmm_2024, fu_mmsm_2021mm} that select keyframes using top-N inter-frame distance, DPSelect identifies additional frames with peak distance as pivot frames, hence the name ``Dist Peak Select''. This strategy is more closely aligned with human temporal perception which tracks the peak stimulus to detect motion \cite{lu_visual_motion_perception_1995}.
Building on DPSelect, to reduce high-level knowledge redundancy, we develop PivotKV to compress KV cache of video sequences in VideoLLMs. Unlike existing LLM compression techniques that treat all tokens equally \cite{zhang_h2o_2023, li_snapkv_2024}, PivotKV preserves pivot frames selected by DPSelect, ensuring critical low-level details remain intact. It then compresses non-pivot visual tokens based on their attention scores, which implicitly capture token redundancy identified by the VideoLLM's high-level multimodal knowledge.

We conduct extensive experiments on various long-video understanding benchmarks, including VideoMME~\cite{fu_videomme_2024}, MLVU~\cite{zhou_mlvu_2024}, LongVideoBench~\cite{wu_longvideobench_2024}, and LVBench~\cite{wang_lvbench_2024}.
Results show that {\RETAKE} enables VideoLLMs to process 8× more frames (up to 2048), allowing them to achieve a 3\%–5\% performance gain over models of similar size.. It even surpasses much larger models, such as VideoLLaMA2-72B~\cite{gvlab_internvl2blog_2024}, LLaVA-OneVision-72B~\cite{cheng_videoLLaMA2_2024}, and Oryx-1.5-34B~\cite{liu_oryx_2024}, on VideoMME, MLVU, and LVBench, respectively.
Additionally, {\RETAKE} introduces only 10\% prefilling latency overhead while reducing decoding latency by 20\%.
Ablation study further validates the complementarity of DPSelect and PivotKV, and {\RETAKE}'s superiority against other video compression methods.1
%
In summary, our contributions are threefold:
\begin{itemize}
    \item To our best knowledge, the training-free {\RETAKE}
    is the first to jointly model temporal and knowledge redundancy for long video compression, enabling VideoLLMs to process 8× longer frames (up 2048).
    \item We propose a novel keyframe selection method DPSelect to reduce low-level temporal redundancy, and a novel token compression method PivotKV to reduce high-level knowledge redundancy in long videos.
    \item {\RETAKE} introduces only 10\% prefilling latency overhead while reducing decoding latency by 20\%. It helps existing VideoLLMs outperform similar-sized VideoLLMs by 3\%-5\%, setting new state-of-the-art benchmarks.
\end{itemize}

\section{Related Work}

\subsection{Video Large Language Models}


Most VideoLLMs build on image MLLMs and are further trained on video-text pairs to capture temporal relationships. A VideoLLM typically comprises a vision foundational model (VFM)~\cite{Radford_clip_2021}, an LLM~\cite{chiang_vicuna_2023}, and a connector linking them. The VFM extracts visual features, while the connector maps them into a format the LLM can process for video understanding.  
VideoLLMs fall into two main types based on the connector: concatenation-based and Q-Former-based. Concatenation-based models~\cite{maaz2023video, lin2023videollava, liu_llava_next_2024} are simple and effective but suffer from exponentially increasing computational costs, limiting their scalability for long videos. Q-Former-based models~\cite{2023videochat, Li_VideoChat2_2024} use a transformer decoder with learnable tokens to compress VFM embeddings, reducing computation but losing information, which impacts performance.

\subsection{Long Video Understanding}

The VideoLLM community has explored three main approaches for long video understanding: 
(1) \textit{Sparse Sampling}: Early VideoLLMs~\cite{zhou_mlvu_2024, wu_longvideobench_2024, fu_videomme_2024} handle long videos by uniformly sampling frames. For instance, Video-LLaVA~\cite{lin2023videollava} and VideoChat2~\cite{Li_VideoChat2_2024} sample 8 and 16 frames, respectively. However, as video length increases, more frames are discarded, disrupting temporal structures~\cite{zhang_longva_2024}.  
(2) \textit{Length Extrapolation}: Recent works~\cite{zhang_longva_2024, shu_video-xl_2024, xue2024longvila} extend VideoLLMs' context length by post-training on long-sequence data, improving performance when more frames are sampled. However, this approach cannot overcome the increasing memory cost brought by longer video sequences beyond 512 frames~\cite{zhang_longva_2024}.  
(3) \textit{Multimodal Token Compression}: To fit more frames within memory limits, \citeauthor{weng_longvlm_2024} and \citeauthor{li2023llama_vid}~\cite{weng_longvlm_2024, li2023llama_vid} reduce token counts by averaging adjacent temporal tokens. \citeauthor{jin_chat-univi_2023} and \citeauthor{ren2023testa}~\cite{jin_chat-univi_2023, ren2023testa} further refine video representations by merging non-essential visual tokens within and across frames. \citeauthor{song2023moviechat} and \citeauthor{he_ma-lmm_2024}~\cite{song2023moviechat, he_ma-lmm_2024} use a memory bank to iteratively merge and store tokens. These methods mainly leverage low-level temporal redundancy between frames, or train a Q-Former with a fixed compression ratio, ignoring the high-level world knowledge within LLM that can be used for compression.

In summary, to perceive more frames given a fixed memory budget, token compression is an effective method. However, existing methods overlook knowledge redundancy, limiting their compression ability. We addresses this by managing both low- and high-level redundancies.
\section{Method}

\subsection{Overview}

\begin{figure}[t]
    \centering
    \includegraphics[width=\linewidth]{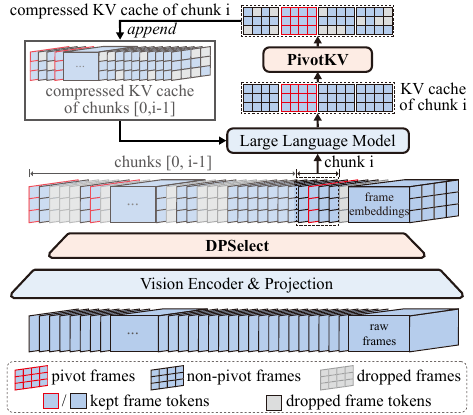}
    \caption{    
    Illustration of ReTaKe.
    DPSelect select keyframes.
    Video sequence is then processed chunk by chunk, during which PivotKV compresses the KV cache of video tokens.
    }
    \label{fig:retake_overview}
\end{figure}

The overall architecture of {\RETAKE} is shown in \autoref{fig:retake_overview}. 
To reduce temporal redundancy before feeding video into LLM, DPSelect extracts keyframes by preserving frames with high inter-frame distances, and marking peak points as pivot frames. 
To reduce memory cost, we then divide the video into smaller chunks of equal length, which are then sequentially prefilled and compressed. Specifically, after prefill a chunk, PivotKV compresses its KV cache to reduce knowledge redundancy: pivot frames remain uncompressed, while non-pivot frames are compressed.

Algorithm~\ref{alg:retake} outlines the detailed pipeline with four steps:
(1) DPSelect frame compression. Raw frames are encoded by a visual encoder, and keyframes (with a pivot mask) are selected using DPSelect (\autoref{sec:dpselect}).
(2) Chunked prefilling. The video sequence is divided into fixed-size chunks, which are processed sequentially (\autoref{sec:preliminary}).
(3) PivotKV token compression. During chunked prefilling, PivotKV (\autoref{sec:pivotkv}) compresses tokens in the video KVCache.
(4) Decoding. Same as general VideoLLMs.


\subsection{Preliminaries: Chunked Prefill} \label{sec:preliminary}

Similar to LLMs, VideoLLMs compute self-attention auto-regressively during the prefill stage, with each token's representation depending only on preceding tokens. Thus, applying \textbf{chunked prefill}~\cite{zeng_chunked_prefill_2024}, which processes input tokens in sequential chunks, remains mathematically equivalent.

\begin{algorithm}[t]
\caption{{\RETAKE} Algorithm} \label{alg:retake}
\begin{algorithmic}[1]
\STATE \textbf{Input:} Video frames $\mathbf{F} \in \mathbb{R}^{T \times 3}$, prompt embeddings $\mathbf{P} \in \mathbb{R}^{L\times d}$, chunk size $\tau$, DPSelect compression ratio $\alpha_{\text{dp}} \in (0,1]$, and PivotKV compression ratio $\alpha_{\text{kv}} \in (0,1]$. Visual encoder with projection $\text{VE}$, visual-text concatenator $\text{VTC}$, and large language model $\text{LLM}$.
\STATE \textbf{Output:} Generated output $O$.

\STATE // Step 1: DPSelect frame compression 
\STATE $\mathbf{M} \in \mathbb{R}^{T \times N \times d} \gets \text{VE}\left( \mathbf{F} \right)$

\STATE $\hat{\mathbf{M}} \in \mathbb{R}^{\alpha_{\text{dp}}T N \times d}, \mathbf{S}\in \{0,1\}^{\alpha_{\text{dp}}T N} \gets \text{DPSelect}\left( \mathbf{M} \right)$

\STATE $\mathbf{H} \in \mathbb{R}^{\left(\alpha_{\text{dp}} T N + L\right) \times d} \gets \text{VTC}\left( \mathbf{P}, \hat{\mathbf{M}} \right)$
\STATE // Step 2: Chunked prefilling
\STATE $\mathcal{H}=\left[ \mathbf{H}_1, \mathbf{H}_2, ..., \mathbf{H}_{\alpha_{\text{dp}} T/\tau+1} \right] \gets \mathbf{H}$

\STATE Initialize KVCache $\mathbf{KV} \in \mathbb{R}^{2\times h\times l\times d_h}$, length $l\gets 0$.
\FOR{$\mathbf{H}_i$ in $\mathcal{H}$}
    \STATE $\mathbf{KV}_i \gets \text{LLM} \left( \mathbf{H}_i, \mathbf{KV} \right) $
    \IF{$\mathbf{H}_i$ is video chunk}
        \STATE // Step 3: PivotKV token compression
        \STATE $\mathbf{KV} \gets \text{PivotKV} \left( \mathbf{KV}, \mathbf{KV}_i, \mathbf{S}\right) $
        \STATE $l\gets l + \alpha_{\text{kv}} \tau N$
    \ELSE
        \STATE $\mathbf{KV} \gets \text{Concat} \left( \mathbf{KV}, \mathbf{KV}_i\right) $
        \STATE $l\gets l + L$
    \ENDIF
\ENDFOR

\STATE // Step 4: Decoding
\STATE $O \gets \text{LLM} \left( \mathbf{K} \right)$
\RETURN{$O$}
\end{algorithmic}
\end{algorithm}


\subsection{DPSelect} \label{sec:dpselect}

\begin{figure}[t]
    \centering
    \includegraphics[width=0.8\linewidth]{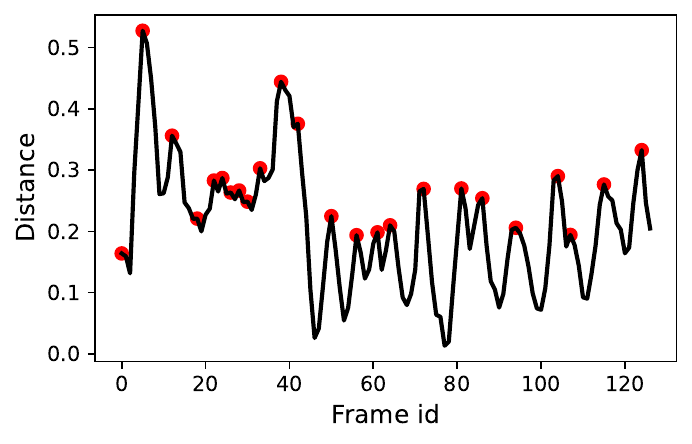}
    \caption{    
    An example of DPSelect. Distance represents cosine dissimilarity between the $i$-th and $i+1$-th frames.
    }
    \label{fig:dpselect_illustration}
\end{figure}

To reduce temporal redundancy, DPSelect extracts keyframes from video features, as shown in \autoref{fig:dpselect_illustration}. Given compression ratio $\alpha_{dp}\in(0,1]$ and an input video feature sequence $\mathbf{M} \in \mathbb{R}^{T \times N \times d}$, where $T$ is the number of frames, $N$ is the number of tokens per frame, and $d$ is the hidden state dimension, DPSelect compresses it into a shorter sequence $\hat{\mathbf{M}} \in \mathbb{R}^{\alpha_{dp}TN \times d}$ and a binary mask indicating pivot tokens $\mathbf{S} \in \{0,1\}^{\alpha_{dp}TN}$.

First, we compute the token-averaged cosine distance between adjacent frames $\textbf{d}\in\mathbb{R}^{T-1}$:
\begin{equation}
\textbf{d} = \frac{1}{N} \sum_{j=1}^{N} \left( 1 - \cos(\mathbf{M}[\text{:-1},j], \mathbf{M}[\text{1:},j]) \right),
\end{equation}
where $[:,:]$ denote matrix indexing operations.
Next, max pooling identifies local maxima in the distance sequence $\textbf{d}$ as pivot frames $\mathcal{P}$, which is closely aligned with human video perception:
\begin{equation}
    \left\{
    \begin{aligned}
        \mathcal{P} &= \arg\max_{w_i} ( \mathbf{d}[w_i]), \\
        w_i &= [i-\lfloor w/2 \rfloor, i+\lfloor w/2 \rfloor],
    \end{aligned}
    \right.
\end{equation}
where $w=3$ is the window size of max pooling. 
Finally, we determine keyframes timestamps $\mathcal{K}$ by selecting local maxima $\mathcal{P}$ and top frames based on cosine distance:
\begin{equation}
\mathcal{K} = \mathcal{P} \cup \text{ArgTop-k}\left( \mathbf{d}[\neg\mathcal{P}], k=\alpha_{\text{dp}}T-|\mathcal{P}| \right).
\end{equation}
The compressed video features $\hat{\mathbf{M}} \in \mathbb{R}^{\alpha_{\text{dp}}T N \times d}$ are extracted from $\mathbf{M}$ based on $\mathcal{K}$ and flattened:
\begin{equation}
    \hat{\mathbf{M}} = \text{Flatten}\left(
        \mathbf{M}[\mathcal{K},:,:]
    \right).
\end{equation}
A binary mask $\mathbf{S} \in \{0,1\}^{\alpha_{\text{dp}}T N}$ is also generated to indicate whether a token in $\hat{\mathbf{M}}$ originates from pivot frames $\mathcal{P}$.




\subsection{PivotKV} \label{sec:pivotkv}

After pre-filling the input video features for the $i$-th chunk, $\mathbf{H}_i \in \mathbb{R}^{\tau N \times d}$, we compute its KV cache for each layer: $\mathbf{K}_i, \mathbf{V}_i \in \mathbb{R}^{\tau N \times d}$, where $\tau$ is the number of frames per chunk. By setting a compression ratio $\alpha_{\text{kv}} \in (0, 1]$, PivotKV compresses the sequence into a shorter form, $\hat{\mathbf{K}}_i, \hat{\mathbf{V}}_i \in \mathbb{R}^{\alpha_{\text{kv}} \tau N \times d}$, while preserving pivot tokens and selecting important tokens based on the attention matrix, which captures redundancy through the MLLM's high-level multimodal knowledge. We omit the layer index since the operations are identical across all layers.

Let $\mathbf{Q}_i \in \mathbb{R}^{h \times l_q \times d_h}$ represent the query states and $\mathbf{K}, \mathbf{V} \in \mathbb{R}^{h \times l_{i-1} \times d_h}$ be the historical KV cache, where $h$ is the number of attention heads, $d_h = d / h$, $l_q = \tau N$ is the chunk length, and $l_{i-1}$ is the length of the historical cache.

First, we compute the attention weights within the current chunk, $\mathbf{A}_i \in \mathbb{R}^{h \times l_q \times l_q}$:
\begin{equation}
    \mathbf{A} = \text{Softmax} \left( \frac{\mathbf{Q} \mathbf{K}_i^\top}{\sqrt{d_h}} \right).
\end{equation}
Next, we compute the importance scores of each token, $\bar{\mathbf{a}} \in \mathbb{R}^{l_q}$, by summing the head-mean attention weights across all keys:
\begin{equation}
    \bar{\mathbf{a}} = \sum_{j=1}^{l_q} \frac{1}{h} \sum_{k=1}^{h} \mathbf{A}_{k,:,j}.
\end{equation}
To mark the pivot tokens, we add the pivot mask $\mathbf{s} \in \{0,1\}^{l_q=\tau N}$ (sliced and flattened from $\mathbf{S}$) for current chunk:
\begin{equation}\label{eq:pivot_masking}
\left\{
\begin{aligned}
    \mathbf{s} &= \mathbf{S}[i \tau N : (i+1) \tau N], \\
    \bar{\mathbf{a}} &= \bar{\mathbf{a}} + \mathbf{s} \cdot \infty.
\end{aligned}
\right.
\end{equation}
Finally, we select the top $\alpha_{\text{kv}} l_q$ importance scores to form the new KV cache by selecting indices $\mathcal{I} \in \mathbb{Z}^K$:
\begin{equation}
\left\{
\begin{aligned}
    \mathcal{I} &= \text{ArgTop-k}(\bar{\mathbf{a}}, k = \alpha_{\text{kv}} l_q), \\
    \hat{\mathbf{K}}_i &= \mathbf{K}_i[:, \mathcal{I},:], \\
    \hat{\mathbf{V}}_i &= \mathbf{V}_i[:, \mathcal{I},:].
\end{aligned}
\right.
\end{equation}
These are then used to update the historical KV cache:
\begin{equation}
\left\{
\begin{aligned}
    \mathbf{K} &\gets \text{Concatenate} (\mathbf{K} || \hat{\mathbf{K}}_i), \\
    \mathbf{V} &\gets \text{Concatenate} (\mathbf{V} || \hat{\mathbf{V}}_i).
\end{aligned}
\right.
\end{equation}

\subsection{Efficiency Optimization} \label{sec:optimize_prefill}

\begin{figure}[t]
    \centering
    \includegraphics[width=\linewidth]{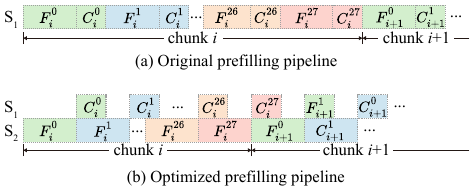}
    \caption{    
    Efficiency optimization through overlapping compression operations with prefilling.
    $\text{S}_1, \text{S}_2$ represent different CUDA streams.
    $F_i^l,C_i^l$ denote prefilling and compression operations, respectively, for chunk $i$ in the $l$-th layer.
    }
    \label{fig:overlap_compression}
\end{figure}


{\RETAKE} introduces additional computational overhead to VideoLLM, primarily within the PivotKV module of the LLM. To optimize this, we incorporate an extra CUDA stream to overlap compression operations, as shown in \autoref{fig:overlap_compression}. Specifically, since PivotKV compression occurs after prefilling in each layer and its cost is lower than prefilling, the compression of layer \( l \) can run simultaneously with the prefilling of layer \( l+1 \).
\section{Experiments}

\begin{table}[t]
\centering

\setlength{\tabcolsep}{2pt}
\resizebox{\linewidth}{!}{
\begin{tabular}{cccccccccc}
\hline
& & \multicolumn{2}{c}{\rotatebox{90}{VideoMME}} & \textbf{} & \rotatebox{90}{MLVU} & \textbf{} & \rotatebox{90}{LongVideoBench} & \textbf{} & \rotatebox{90}{LVBench} \\ \cline{3-4} \cline{6-6} \cline{8-8} \cline{10-10} 
\multirow{-2}{*}{Methods} & \multirow{-2}{*}{\textbf{}} & Long & Overall &  & Dev & {\color[HTML]{363636} } & {\color[HTML]{363636} Val} & {\color[HTML]{363636} } & {\color[HTML]{363636} Val} \\ \hline
\rowcolor[HTML]{F2F2F2} 
GPT-4o~\cite{openai2024gpt4o} &  & 65.3 & 71.9 &  & 64.6 &  & 66.7 &  & 30.8 \\
\rowcolor[HTML]{F2F2F2} 
Gemini 1.5 Pro~\cite{Reid_Gemini1.5_2024} &  & 67.4 & 75.0 &  & - &  & 64.0 &  & 33.1 \\ \hline
MovieChat-7B~\cite{song2023moviechat} &  & - & - &  & 25.8 &  & - &  & 22.5 \\
LLaMA-VID-7B~\cite{li2023llama_vid} &  & - & - &  & 33.2 &  & - &  & 23.9 \\
MA-LMM-7B~\cite{he_ma-lmm_2024} &  & - & - &  & 36.4 &  & - &  & - \\
VideoChat2-7B~\cite{Li_VideoChat2_2024} &  & 33.2 & 39.5 &  & 44.5 &  & 39.3 &  & - \\
Video-LLaVA-7B~\cite{lin2023videollava} &  & 36.2 & 39.9 &  & 47.3 &  & - &  & - \\
{\color[HTML]{BFBFBF} InternVL-1.5-20B~\cite{Chen_InternVL_2024}} & {\color[HTML]{BFBFBF} } & {\color[HTML]{BFBFBF} 45.6} & {\color[HTML]{BFBFBF} 50.7} & {\color[HTML]{BFBFBF} } & {\color[HTML]{BFBFBF} 50.4} &  & {\color[HTML]{BFBFBF} 51.2} & {\color[HTML]{BFBFBF} } & {\color[HTML]{BFBFBF} 39.6} \\
LongVILA-8B~\cite{xue2024longvila} &  & 39.7 & 50.5 &  & 56.7 &  & 57.7 &  & - \\
LongVA-7B~\cite{zhang_longva_2024} &  & 46.2 & 52.6 &  & 63.5 &  & - &  & - \\
mPLUG-Owl3-8B~\cite{ye_mplug_owl3_2025} &  & - & 53.5 &  & 63.7 &  & 52.1 &  & - \\
LLaVA-Octopus-7B~\cite{zhao_llava_octopus_2025} &  & - & 54.7 &  & 57.5 &  & - &  & - \\
VITA-1.5-7B~\cite{fu_vita-15_2025} &  & - & 56.1 &  & - &  & - &  & - \\
LongViTU-7B~\cite{wu_longvitu_2025} &  & 48.4 & 56.3 &  & - &  & - &  & - \\
TimeMarker-8B~\cite{github_timemarker_2024} &  & - & 57.3 &  & 63.9 &  & 56.3 &  & 41.3 \\
Video-XL-7B~\cite{shu_video-xl_2024} &  & 49.2 & 55.5 &  & 64.9 &  & - &  & - \\
{\color[HTML]{BFBFBF} Video-LLaMA2-72B~\cite{cheng_videoLLaMA2_2024}} & {\color[HTML]{BFBFBF} } & {\color[HTML]{BFBFBF} 57.6} & {\color[HTML]{BFBFBF} 62.4} & {\color[HTML]{BFBFBF} } & {\color[HTML]{BFBFBF} 61.2} &  & {\color[HTML]{BFBFBF} -} & {\color[HTML]{BFBFBF} } & {\color[HTML]{BFBFBF} -} \\
\rowcolor[HTML]{EAFAF1} 
Qwen2-VL-7B~\cite{wang_qwen2-vl_2024} &  & 53.8 & 63.3 &  & 64.8 &  & 55.6 &  & 42.4 \\
\rowcolor[HTML]{EAFAF1} 
\textbf{QWen2VL-7B+ReTaKe} &  & \textbf{56.2} & 63.9 & \textbf{} & \textbf{69.8} & \textbf{} & 57.7 & \textbf{} & 47.8 \\
\rowcolor[HTML]{EAFAF1} 
LLaVA-Video-7B~\cite{zhang_llava-video_2024} &  & 52.3 & 62.2 &  & 64.9 &  & 58.2 &  & 43.3 \\
\rowcolor[HTML]{EAFAF1} 
\textbf{LLaVA-Video-7B+ReTaKe} &  & 55.4 & \textbf{64.0} & \textbf{} & 67.8 & \textbf{} & \textbf{59.7} & \textbf{} & \textbf{48.5} \\
{\color[HTML]{BFBFBF} LLaVA-Onevision-72B~\cite{li_llava-onevision_2024}} & {\color[HTML]{BFBFBF} } & {\color[HTML]{BFBFBF} 60.0} & {\color[HTML]{BFBFBF} 66.3} & {\color[HTML]{BFBFBF} } & {\color[HTML]{BFBFBF} 66.4} &  & {\color[HTML]{BFBFBF} 61.3} & {\color[HTML]{BFBFBF} } & {\color[HTML]{BFBFBF} -} \\
{\color[HTML]{BFBFBF} Aria-25B~\cite{li_aria_2024}} & {\color[HTML]{BFBFBF} } & {\color[HTML]{BFBFBF} 59.3} & {\color[HTML]{BFBFBF} 67.6} & {\color[HTML]{BFBFBF} } & {\color[HTML]{BFBFBF} 70.6} & \multicolumn{1}{l}{} & {\color[HTML]{BFBFBF} 64.2} & {\color[HTML]{BFBFBF} } & {\color[HTML]{BFBFBF} -} \\
{\color[HTML]{BFBFBF} Oryx-1.5-34B~\cite{liu_oryx_2024}} & {\color[HTML]{BFBFBF} } & {\color[HTML]{BFBFBF} 58.8} & {\color[HTML]{BFBFBF} 67.3} & {\color[HTML]{BFBFBF} } & {\color[HTML]{BFBFBF} 72.3} &  & {\color[HTML]{BFBFBF} 62.0} & {\color[HTML]{BFBFBF} } & {\color[HTML]{BFBFBF} 30.4} \\
{\color[HTML]{BFBFBF} LLaVA-Video-72B~\cite{zhang_llava-video_2024}} & {\color[HTML]{BFBFBF} } & {\color[HTML]{BFBFBF} 61.5} & {\color[HTML]{BFBFBF} 70.5} & {\color[HTML]{BFBFBF} } & {\color[HTML]{BFBFBF} 74.4} &  & {\color[HTML]{BFBFBF} 61.9} & {\color[HTML]{BFBFBF} } & {\color[HTML]{BFBFBF} -} \\ \hline
\end{tabular}
}

\caption{
Performance comparison. 
{\RETAKE} consistently enhances performance on long video understanding benchmarks, with greater gains for longer videos in LVBench.
}
\label{tab:baseline_comp}

\end{table}

\subsection{Benchmarks and Implementations}


\textbf{Video-MME}~\cite{fu_videomme_2024} contains 900 videos and 2,700  Multiple-Choice Question-Answer (MCQA) pairs, categorized into short ($<$2 min), medium, and long (30–60 min) subsets.   
\textbf{MLVU}~\cite{zhou_mlvu_2024} includes videos ranging from 3 minutes to 2 hours and spans 9 tasks.   
\textbf{LongVideoBench}~\cite{wu_longvideobench_2024} comprises 3,763 videos (up to 1 hour) and 6,678 MCQA pairs.   
\textbf{LVBench}~\cite{wang_lvbench_2024} averages 4,101 seconds per video—4 times longer than Video-MME and 5 times longer than MLVU. It includes 1,549 MCQA pairs across 6 tasks. All datasets are human annotated.




\paragraph{Implementation Details.}  
Our experiments are based on QWen2VL-7B~\cite{wang_qwen2-vl_2024} and LLaVA-Video-7B~\cite{zhang_llava-video_2024}, extended with {\RETAKE}.  
Following the original model settings, we resized the longer side of input frames to 448 pixels for QWen2VL and the shorter side to 336 pixels for LLaVA-Video.  
For the default sampling strategy, videos were densely sampled at 2 frames per second (FPS), with a maximum of 2048 frames for QWen2VL and 1024 frames for LLaVA-Video (i.e., for longer videos, FPS was reduced to maintain the number of frames to these limits).  
For maximal context length in default, we adjusted the compression ratio for each video to make sure its context length does not exceed 32,000.
For certain ablation studies, we reduced the maximal sampled frames to 256, as the upper limit for processing frames without compression is approximately 300.  

\subsection{Main Results}\label{mainresult}


\paragraph{Comparison with SoTAs.} We conducted experiments on VideoMME~\cite{fu_videomme_2024}, MLVU~\cite{zhou_mlvu_2024}, LongVideoBench~\cite{wu_longvideobench_2024}, and LVBench~\cite{wang_lvbench_2024}, as shown in \autoref{tab:baseline_comp}. 
For base VideoLLMs with different architectures, {\RETAKE} consistently improves performance on long video understanding, achieving an average gain of 3.1\% and 2.9\% for QWen2VL-7B~\cite{wang_qwen2-vl_2024} and LLaVA-Video-7B~\cite{zhang_llava-video_2024}, respectively. Notably, for LVBench, which features the longest average video length, the performance gain reaches a remarkable 5.3\%.
Additionally, {\RETAKE} outperforms existing models of similar size, including LongVA~\cite{zhang_longva_2024}, LongVILA-7B~\cite{xue2024longvila}, and LLaVA-OneVision-7B~\cite{li_llava-onevision_2024}, with improvements ranging from 1.6\% to 4.9\%. It even surpasses much larger models, such as VideoLLaMA2-72B~\cite{cheng_videoLLaMA2_2024} on VideoMME, LLaVA-OneVision-72B~\cite{li_llava-onevision_2024} on MLVU, and Oryx-1.5-34B~\cite{liu_oryx_2024} on LVBench. Furthermore, {\RETAKE} significantly outperforms GPT-4o~\cite{openai2024gpt4o} on both MLVU and LVBench.
These benchmarks cover diverse video durations and question types, demonstrating the robustness and generality of {\RETAKE}.

\begin{table}[t]

\centering
\resizebox{\linewidth}{!}{
\begin{tabular}{llll}
\hline
Methods & FLOPs (T) & TTFT (s) & TPOT (ms) \\ \hline
QWen2VL-7B & 3.0 & 5.7 & 116.7 \\
+{\RETAKE} & 2.7 \footnotesize{\textcolor{red}{-9\%}} & 7.4 \footnotesize{\textcolor{green}{+28\%}} & 93.9 \footnotesize{\textcolor{red}{-20\%}} \\
+{\RETAKE}-OPT & 2.7 \footnotesize{\textcolor{red}{-9\%}} & 6.2 \footnotesize{\textcolor{green}{+8\%}} & 94.2 \footnotesize{\textcolor{red}{-19\%}} \\ \hline
LLaVA-Video-7B & 10.4 & 4.4 & 58.5 \\
+{\RETAKE} & 8.6 \footnotesize{\textcolor{red}{-18\%}} & 7.2 \footnotesize{\textcolor{green}{+62\%}} & 42.7 \footnotesize{\textcolor{red}{-27\%}} \\
+{\RETAKE}-OPT & 8.6 \footnotesize{\textcolor{red}{-18\%}} & 4.9 \footnotesize{\textcolor{green}{+11\%}} & 42.3 \footnotesize{\textcolor{red}{-28\%}} \\ \hline
\end{tabular}
}

\caption{
Efficiency analysis.
TTFT and TPOT denote Time To First Token and Time Per Output Token.
With overlapping compression optimization, {\RETAKE} significantly reduces decoding latency (TPOT) with slight increase in prefilling latency (TTFT).
}
\label{tab:efficiency}

\end{table}

\paragraph{Efficiency Analysis.}

\begin{table*}[t!]

\centering

\setlength{\tabcolsep}{4pt}
\resizebox{\textwidth}{!}{
\begin{tabular}{lccccccc}
\hline
\multicolumn{1}{c}{{\color[HTML]{363636} Model}} & {\color[HTML]{363636} Max Frames} & {\color[HTML]{363636} Max Context Length} & {\color[HTML]{363636} VideoMME-L} & {\color[HTML]{363636} MLVU} & {\color[HTML]{363636} LongVideoBench} & {\color[HTML]{363636} LVBench} & $\Delta_{avg}$ \\ \hline
0 QWen2VL-7B & 256 & 18K & 53.8 & 64.8 & 55.6 & 42.6 & - \\
\rowcolor[HTML]{F2F2F2} 
1 +scale up frames directly & 512 & 36K & \multicolumn{4}{c}{\cellcolor[HTML]{F2F2F2}Out of Memory} & - \\
2 +{\RETAKE} compression & 256 & 16K & 52.7 & 64.7 & 55.1 & 42.3 & \textcolor{red}{-0.5} \\
3 +scale up frames 2x & 512 & 16K & 53.6 & 68.6 & 56.8 & 44.8 & \textcolor{green}{+2.3} \\
4 +scale up frames 4x & 2048 & 16K & \textbf{56.2} & 69.0 & 56.3 & 46.3 & \textcolor{green}{+1.0} \\
5 +scale up context length & 2048 & 32K & 55.7 & \textbf{69.8} & \textbf{57.7} & \textbf{47.8} & \textcolor{green}{+0.8} \\ \hline
\end{tabular}
}

\caption{
    In-depth analysis for more insights. By compressing the videos of varying length into a fixed context length, {\RETAKE} allows the model to process more frames under fixed memory budget, thereby achieving significant gains.
}
\label{tab:indepth_analysis}

\end{table*}

\begin{figure*}[t]
    \centering
    \includegraphics[width=\linewidth]{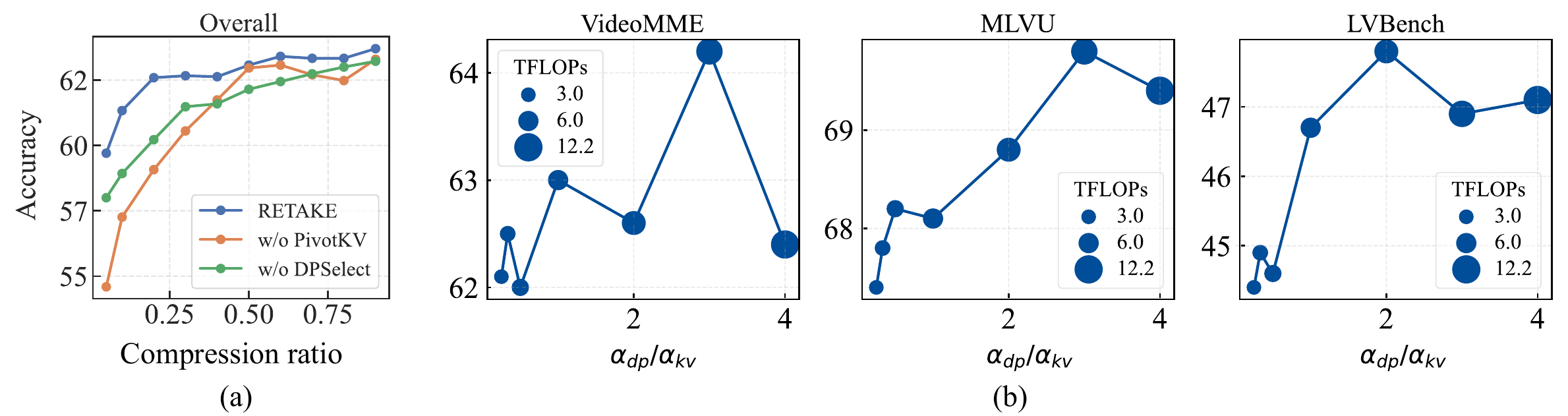}
    \caption{ 
    (a) Ablation study on DPSelect and PivotKV under different compression ratios.  
    (b) Trade-off between knowledge and temporal redundancy—{\RETAKE} favors leveraging knowledge redundancy.  
    }
    \label{fig:ablation_studies}
\end{figure*}

To evaluate the efficiency of {\RETAKE}, we measured its FLOPs and TTFT for prefilling and TPOT for decoding with 256 max frames and 0.5 compression ratio, shown in \autoref{tab:efficiency}. 
{\RETAKE} significantly reduces both FLOPs and TPOT, achieving a 19\% TPOT and 8\% FLOPs reduction for QWen2VL-7B, and a 26\% TPOT and 17\% FLOPs reduction for LLaVA-Video-7B. These improvements stem from {\RETAKE}'s ability to compress the video sequence length, lowering both prefilling computational cost (FLOPs) and decoding latency (TPOT). 
However, without optimization, {\RETAKE} incurs notable overhead (+28\% TTFTs for QWen2VL and +62\% for LLaVA-Video), which the optimized version ({\RETAKE}-OPT) significantly reduces to just 8\% and 11\%, demonstrating the efficiency of our implementation.

\subsection{In-depth Analysis}


\paragraph{Insights on how {\RETAKE} Improves Performance.}  
To investigate the sources of performance gains, we conducted ablation studies summarized in \autoref{tab:indepth_analysis}.  
\#0 serves as the baseline model without modifications.  
In \#1, directly increasing the number of frames exceeded the A100 GPU's memory limit. For long-context inference, memory consumption is primarily determined by the KVCache context length~\cite{hooper_kvquant_2024}.
Thus, in \#2, we compressed videos to a maximum context length of 16K tokens—compressing longer videos while leaving shorter ones unchanged—resulting in a 0.5\% average accuracy drop.  
With this fixed budget, {\RETAKE} compression enables scaling up the number of frames in \#3 and \#4, improving accuracy by 2.3\% and 1.0\% on average. Notably, for MLVU and LongVideoBench, increasing frames from 512 to 2048 leads to fluctuations, likely due to saturation: these datasets contain relatively short videos (e.g., MLVU averages 10 minutes), meaning that under a fixed FPS, the sampled frames rarely reach the upper limit.  
In contrast, for longer videos like VideoMME-L and LVBench, performance improvements remained significant.  
Finally, expanding the maximum context length to 32K further enhanced performance in \#5.

\subsection{Ablation Studies}\label{ablation}

\paragraph{Ablation of DPSelect and PivotKV.}
We evaluated two variants in QWen2VL: \textbf{w/o DPSelect}, where PivotKV ignores pivot tokens in \autoref{eq:pivot_masking}, and \textbf{w/o PivotKV}, where only DPSelect remains active. 
To study their behavior under different compression ratios, we set DPSelect compression ratio $\alpha_{dp}=1$ (only provide pivots tokens), and adjusted the PivotKV compression ratio $\alpha_{kv}$ for each video, results are shown in \autoref{fig:ablation_studies} (a).

The following observations can be made:  
(1) {\RETAKE} outperforms both ``w/o PivotKV'' and ``w/o DPSelect,'' particularly at lower compression ratios where videos are compressed into short sequences.  
(2) ``w/o PivotKV'' experiences a significantly steeper performance drop than ``w/o DPSelect'' and {\RETAKE}, indicating that solely reducing temporal redundancy leads to substantial information loss. This highlights the necessity of also minimizing knowledge redundancy to maintain performance under the same compression ratio. Consequently, we further explore the trade-off between these two redundancies below.

\begin{table*}[t!]

\centering

\begin{tabular}{cccccclcc}
\hline
 &  &  & \textbf{} & \multicolumn{2}{c}{MLVU} & \multicolumn{1}{c}{\textbf{}} & \multicolumn{2}{c}{LVBench} \\
\multirow{-2}{*}{Methods} & \multirow{-2}{*}{Max Frames} & \multirow{-2}{*}{Max Context Length} &  & NQA & AO & \multicolumn{1}{c}{} & KIR & {\color[HTML]{363636} TG} \\ \hline
LLaVA-Video-7B~\cite{zhang_llava-video_2024} & 128 & 25K &  & 74.2 & 55.6 & \multicolumn{1}{c}{} & 37.5 & 36.8 \\
LLaVA-Video-7B+ReTaKe & 1024 & 16K &  & 74.2 & 60.6 & \multicolumn{1}{c}{} & 51.2 & 43.2 \\ \hline
Qwen2-VL-7B~\cite{wang_qwen2-vl_2024} & 256 & 18K &  & 81.9 & 49.0 &  & 44.3 & 40.5 \\
QWen2VL-7B+ReTaKe & 256 & 16K &  & 81.0 & 49.4 &  & 44.7 & 41.4 \\
QWen2VL-7B+ReTaKe & 512 & 16K &  & 81.3 & 59.5 &  & 47.7 & 42.3 \\
QWen2VL-7B+ReTaKe & 1024 & 16K &  & 82.7 & 59.5 &  & 49.5 & 44.1 \\ \hline
\end{tabular}

\caption{
    Investigation of {\RETAKE}'s fine-grained temporal perception capabilities across related task types in MLVU and LVBench datasets: Needle QA (NQA), Action Order (AO), Action Count (AC), Key Information Retrieval (KIR), and Temporal Grounding (TG).
    While {\RETAKE} compression harms needle test performance, it slightly improves other tasks. 
    Moreover, with frame scaling enabled by {\RETAKE}, the model not only compensates for this loss but even surpasses its original performance in fine-grained temporal perception tasks.
}
\label{tab:temporal_fine}

\end{table*}

\paragraph{Trade-off between Knowledge and Temporal Redundancy.}  
We analyzed this trade-off based on QWen2VL in \autoref{fig:ablation_studies} (b) by fixing the total compression rate at $\alpha_{\text{dp}}\alpha_{\text{kv}}=0.25$ and varying $\alpha_{\text{dp}}/\alpha_{\text{kv}}$ to control the balance.  
Since the compression ratio represents the ratio of context length after to before compression, a higher $\alpha_{\text{dp}}/\alpha_{\text{kv}}$ suggests a greater reliance on knowledge redundancy for compression.
Optimal performance is achieved when $\alpha_{\text{dp}}/\alpha_{\text{kv}}$ is between 2 and 3, suggesting a preference for leveraging knowledge redundancy.  
However, increasing $\alpha_{\text{dp}}/\alpha_{\text{kv}}$ also raises FLOPs, as DPSelect reduces the number of visual tokens processed by the LLM, leading to a larger FLOPs reduction at the same compression ratio.  
This reveals an additional trade-off between performance and efficiency.



\paragraph{Investigation on Fine-grained Temporal Perception Abilities.} 
To evaluate the impact of token compression algorithms on critical temporal details, we conducted ablation studies on the MLVU and LVBench datasets.
We compared the baseline models LLaVA-Video-7B~\cite{zhang_llava-video_2024} and QWen2VL-7B~\cite{wang_qwen2-vl_2024}. The results are presented in \autoref{tab:temporal_fine}.

Our analysis yields several key findings:
(1) Comparing two QWen2VL experiments with 256 frames, token compression in {\RETAKE} slightly reduced Needle QA performance from 81.9 to 81.0, which aligns with expectations, as compression inevitably leads to some information loss.
(2) For other tasks, {\RETAKE} compression resulted in slight performance gains. This is reasonable, as eliminating redundant visual information can enhance performance in relatively coarse-grained tasks~\cite{wang_rtq_2023, chen_fastv_2024}.
(3) Since {\RETAKE} enables the model to process more frames, we observed that such frame scaling compensates for the loss in Needle QA for both LLaVA-Video and QWen2VL. Moreover, it significantly improves performance in fine-grained temporal perception tasks, including Action Order (+7.8\%), Key Information Retrieval (+9.5\%), and Temporal Grounding (+5.0\%).
(4) The improvement in MLVU’s Action Order category was notably higher than in Needle QA (7.8\% vs. 0.4\% on average). We attribute this to {\RETAKE}’s ability to sample more frames through token compression, effectively achieving denser frame sampling, which is known to greatly enhance action understanding~\cite{2023videochat, wang_qwen2-vl_2024}.
(5) In LVBench, Key Information Retrieval exhibited a significantly higher improvement than Temporal Grounding, with average gains of 9.5\% versus 5.0\%. We hypothesize that token compression increases information density, facilitating a more comprehensive understanding. Therefore, Key Information Retrieval, a task requiring deeper comprehension, benefits more than perceptual tasks like Temporal Grounding.

\begin{table}[t]

\centering
\begin{tabular}{ccccc}
\hline
{\color[HTML]{363636} } & \multicolumn{2}{c}{{\color[HTML]{363636} VideoMME}} & {\color[HTML]{363636} MLVU} & {\color[HTML]{363636} LVBench} \\ \cline{2-5} 
\multirow{-2}{*}{{\color[HTML]{363636} Method}} & {\color[HTML]{363636} Long} & {\color[HTML]{363636} Overall} & Val & Val \\ \hline
M$^2$SM~\cite{fu_mmsm_2021mm} & 49.1 & 58.0 & 61.6 & 39.7 \\
A2Summ~\cite{he_a2sum_2023} & 48.2 & 57.6 & 60.9 & 39.8 \\
MA-LLM~\cite{he_ma-lmm_2024} & 50.7 & 59.2 & 62.8 & 40.8 \\
DPSelect & 51.0 & 59.5 & 63.2 & 41.4 \\ \hline
FastV~\cite{chen_fastv_2024} & 53.5 & 61.2 & 63.2 & 42.3 \\
FitPrune~\cite{ye_fitprune_2024} & 53.6 & 61.2 & 63.6 & 42.0 \\
LOOK-M~\cite{wan_look-m_2024} & 53.6 & 61.0 & 63.8 & 42.6 \\
SparseVLM~\cite{zhang_sparsevlm_2024} & 54.4 & 60.7 & 63.0 & 43.9 \\
PyramidDrop~\cite{xing_pyramiddrop_2024} & 53.1 & 60.5 & 63.7 & 41.6 \\
VL-Cache~\cite{tu_vl-cache_2024} & 53.2 & 61.3 & 64.5 & 42.4 \\
{\RETAKE} & 55.6 & 64.1 & 69.4 & 45.7 \\ \hline
\end{tabular}

\caption{
Performance comparison with existing keyframe selection and token compression methods.
}
\label{tab:compression_comp}
\end{table}

\begin{figure*}[t!]
    \centering
    \begin{subfigure}{0.99\textwidth}
        \centering
        \includegraphics[width=0.99\linewidth]{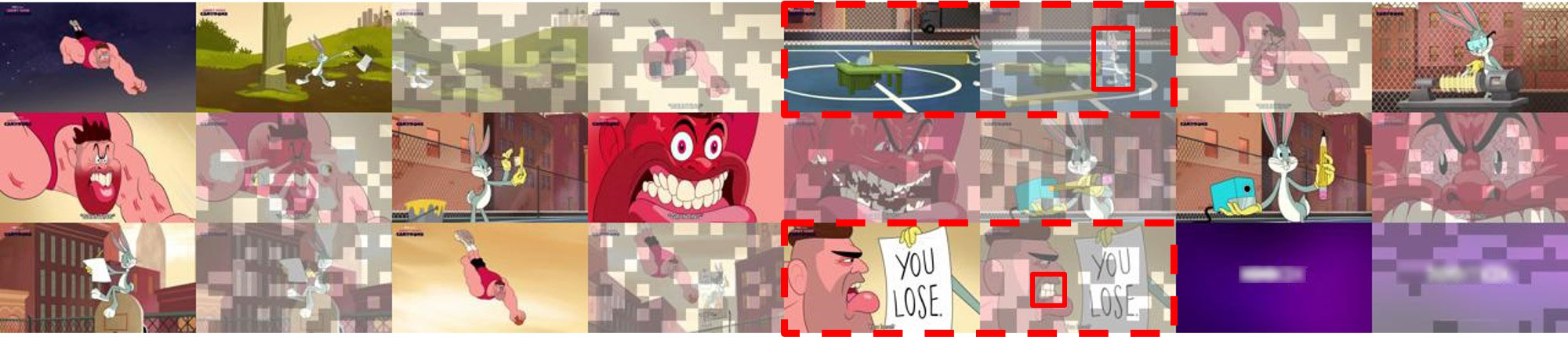}
        \caption{}
        \label{fig:subfig1}
    \end{subfigure}%
    \hfill
    \begin{subfigure}{0.99\textwidth}
        \centering
        \includegraphics[width=0.99\linewidth]{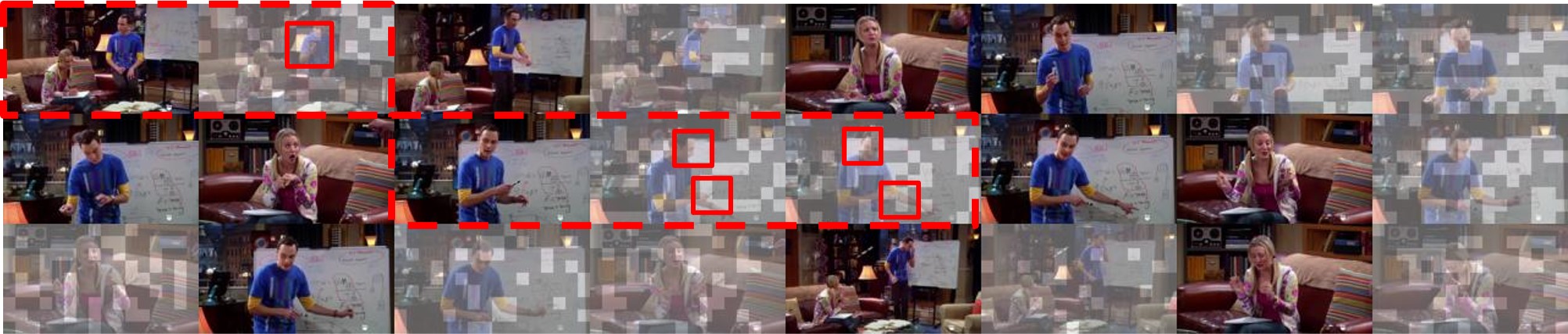}
        \caption{}
        \label{fig:subfig2}
    \end{subfigure}%
    \hfill
    \begin{subfigure}{0.99\textwidth}
        \centering
        \includegraphics[width=\textwidth]{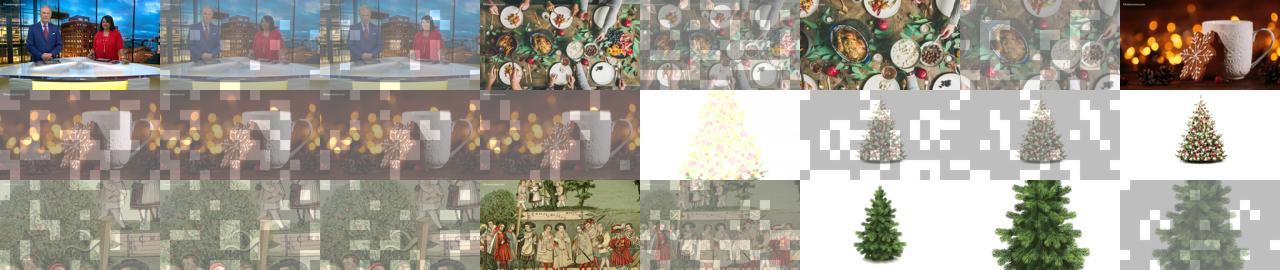}
        \caption{}
        \label{fig:subfig3}
    \end{subfigure}
    \caption{
    Visualization of how {\RETAKE} reduce redundancies in video.
    }
    \label{fig:case_vis}
\end{figure*}

\paragraph{Performance Comparison of Different Compression Methods.}
We evaluated {\RETAKE} against other video compression methods on QWen2VL~\cite{wang_qwen2-vl_2024}. 
For keyframe selection, we set the maximal number of frames to 256 (the common limit of all these models) and compression ratio to 0.5. Our DPSelect consistently outperforms existing baselines. 
Regarding token compression, we set the maximal context length to 16K. {\RETAKE} significantly surpasses previous methods. The reason is that, unlike existing approaches, which primarily compress visual tokens based on prompt tokens—requiring the VideoLLM to process the entire video and prompt sequence—{\RETAKE} exploits inter-visual redundancy to compress tokens in smaller chunks. This strategy substantially increases the maximum number of frames the model can handle from 256 to 2048, enhancing efficiency in long video processing.

\subsection{Visualization}

To qualitatively evaluate the effectiveness of {\RETAKE}, we present some visualization cases in \autoref{fig:case_vis}. The keyframes selected by DPSelect remain unchanged, while non-keyframes are painted white, and patches filtered by PivotKV are painted dark. 
From all cases in \autoref{fig:case_vis}, we can find that DPSelect can always effectively select important frames that differ from adjacent frames, filtering out redundant frames in static scenes. 
Besides, red boxes show that PivotKV can always filter the patches similar to the keyframes and remains the patches with subtle semantic changes between non-keyframes and keyframes, such as body movement and facial expressions. For example, the region where a small animal appears (refer to the first red box of \autoref{fig:subfig1}), and the area where the man's mouth changes (refer to the second red box of \autoref{fig:subfig1}), are identified and left uncompressed, while in red boxes of \autoref{fig:subfig2}, the frequently changing facial and arm regions of the character are selected and preserved. 
This validates our initial motivation that DPSelect reduces low-level temporal redundancy and PivotKV reduces high-level knowledge redundancy. 
However, there are no significant motions in \autoref{fig:subfig3}, and as a result, the filtered token distribution is relatively sparse.

\section{Conclusion}

This paper introduces {\RETAKE}, a training-free approach to understanding long videos by jointly reducing temporal and knowledge redundancy with two novel modules: DPSelect and PivotKV. DPSelect, inspired by human perception, identifies keyframes with significant motion peaks, while PivotKV compresses the KV cache of non-keyframes using knowledge-relevant attention scores.
Experiments demonstrate that {\RETAKE}, with little computational overhead, can handle video sequences up to 8 times longer and thereby achieve state-of-the-art performance.


{
    \small
    \bibliographystyle{ieeenat_fullname}
    \bibliography{main}
}


\end{document}